\documentclass{article}
\usepackage{aistats2016mod}
\usepackage{url}
\usepackage{graphicx}
\usepackage{cite}
\usepackage{amsmath}
\usepackage{amsxtra}
\usepackage{amsfonts}
\usepackage{amssymb}
\usepackage{amsthm}
\usepackage{bm}
\usepackage{mathtools}
\usepackage{float}
\usepackage{subcaption}
\usepackage[commentsnumbered]{algorithm2e}

\newcommand{\pd}[2]{\frac{\partial #1}{\partial #2}}

\newcommand{\mb}{\mathbf}

\newcommand{\argmin}{\operatornamewithlimits{argmin}}

\newcommand{\paren}[1]{\left( #1 \right)}
\newcommand{\bracket}[1]{\left[ #1 \right]}

\newcommand{\set}[1]{\{ #1 \}}

\newcommand{\eps}{\varepsilon}

\begin{document}
\twocolumn[

%% ?? TODO delete me. just for evaluating length while writing
% \cfoot{\thepage}

\aistatstitle{A Markov Jump Process for\\More Efficient Hamiltonian Monte Carlo}

% \aistatsauthor{ Anonymous Author 1 \And Anonymous Author 2 \And Anonymous Author 3 \And Anonymous Author 4}
% \aistatsaddress{ Unknown Institution 1 \And Unknown Institution 2 \And Unknown Institution 3 \And Unknown Institution 4}
\aistatsauthor{ Andrew B Berger \And Mayur Mudigonda  \And Michael R DeWeese \And Jascha Sohl-Dickstein}
\aistatsaddress{ 
    Redwood Center\\UC Berkeley \And
    Redwood Center\\UC Berkeley \And
    Redwood Center\\UC Berkeley \And
    Stanford University\footnotemark
}
% \footnote{Currently at Google Research.}}
% \aistatsaddress{ \And Redwood Center for Theoretical Neuroscience, UC Berkeley \And Stanford University\footnote{Currently at Google Research.}}
]

\newcommand{\fix}{\marginpar{FIX}}
\newcommand{\new}{\marginpar{NEW}}

\begin{abstract}
In most sampling algorithms, including Hamiltonian Monte Carlo, transition rates between states correspond to the probability of making a transition in a single time step, and are constrained to be less than or equal to 1. We derive a Hamiltonian Monte Carlo algorithm using a continuous time Markov jump process, and are thus able to escape this constraint. Transition rates in a Markov jump process need only be non-negative. We demonstrate that the new algorithm leads to improved mixing for several example problems, both by evaluating the spectral gap of the Markov operator, and by computing autocorrelation as a function of compute time. We release the algorithm as an open source Python package.
\end{abstract}

\footnotetext{Currently at Google Research.}

\section{Introduction}

Efficient sampling is a challenge in many tasks involving high dimensional probabilistic models, in a diversity of fields. 
For example, 
sampling is commonly required to train a probabilistic model, to evaluate the model's performance, to perform inference, and to take expectations under the model \cite{MacKay2003a}.

In this paper we introduce a method for more efficient sampling, by making Markov transitions in continuous rather than discrete time. 
This allows transitions into lower probability states to occur more often, with a shorter time spent for each visit, and thus allows for more rapid exploration of state space. 
We apply this approach to develop a novel Hamiltonian Monte Carlo (HMC) sampling algorithm. 
Finally, we demonstrate the effectiveness of this approach by comparing both spectral gaps and autocorrelation on several example problems.

\subsection{Discrete time sampling}

Most sampling algorithms involve transitioning between states in discrete time steps, with a fixed interval.
In this discrete-time framework, the transition rates out of a state must sum to 1 and be non-negative.
In fact, the popular Metropolis-Hastings acceptance rule \cite{Hastings:1970p13277} for Markov Chain Monte Carlo (MCMC) works well because it maximizes the transition rate between a pair of states,
subject to this constraint and to detailed balance. 
As we will see, however, this constraint on transition rates limits performance, and better mixing can be achieved by allowing transition rates larger than 1.

\subsection{Markov jump process}

A Markov process can also be expressed in continuous time, in which case the only restriction on the transition rates between distinct states is that they be non-negative.
In continuous time, the rate of transition from a state $j$ into a state $i\neq j$ is given by $\Gamma_{ij}$, and the rate of change of the probability $p_i$ of state $i$ is
\begin{align}
  \label{eq dpdt}
  \pd{p_i}{t} = \sum_j \Gamma_{ij} p_j ,
\end{align}

where $\Gamma_{ij} \geq 0$ for $\forall i \neq j$, and we use the convention $\Gamma_{jj} = -\sum_{i\neq j} \Gamma_{ij}$.

If a particle evolves in a system of this form it makes stochastic transitions between a set of discrete states
in continuous time.
Each transition is governed by a Poisson process.
Neglecting other states, the waiting time $w_{ij}$ for a transition from state $j$ to state $i$ will be drawn from an exponential distribution, $P(w_{ij}) = \Gamma_{ij}\exp\left( - \Gamma_{ij} w_{ij}\right)$.

To simulate the system, waiting times $w_{kj}$ are generated for all candidate states $k$, and the shortest waiting time $i = \argmin_k w_{kj}$ is chosen. We call this shortest waiting time the holding time. 
A transition is then performed to state $i$ after a delay of length $w_{ij}$.

A system that evolves in this way is known as a Markov jump process \cite{cinlar2013introduction}.
Markov jump processes have been used to model physical systems, such as chemical reactions, which are well described by stochastic transitions between discrete states~\cite{anderson2011continuous}.
Work has also been done on efficient sampling of trajectories in Markov jump processes~\cite{rao2012fast} and the statistical properties of these trajectories~\cite{metzner2009transition}.

To our knowledge, Markov jump processes have not previously been applied to general purpose Monte Carlo sampling, though see \cite{grenander1994} where a Markov jump process is used to sample from a posterior distribution over model graph structure. 
One barrier to using a Markov jump process for general Monte Carlo is that it is necessary to compute transition rates to all possible target states
$i$ from the current state $j$.
As we will see, however, 
for HMC we only need to consider a small number of target states.
\textbf{The primary contribution of this paper is to use a Markov jump process to develop a faster HMC algorithm}.

\subsubsection{Relationship to importance sapling}

As will be elaborated in Section \ref{sec sys comp}, samples from a Markov jump process can be generated by using an equivalent discrete time process $\mb{\hat{T}}$ to generate the same distribution over state \emph{sequences}, and then resampling according to each state's holding time. From this perspective, the process of sampling from a Markov jump process can be seen as a realization of importance sampling, with a particularly unusual proposal distribution. The equivalent discrete time process $\mb{\hat{T}}$ defines the importance sampling proposal distribution, and the holding times provide the importance weights.

The discrete time distribution $\mb{\hat{p}}$ generated by $\mb{\hat{T}}$ will tend to be more similar to a uniform distribution than $\mb p$, and the corresponding Markov chain $\mb{\hat{T}}$ will thus tend to mix more quickly than a typical discrete time sampler. For instance, $\mb{\hat{T}}$ has no self-transitions, so unlike in a standard Metropolis-Hastings algorithm there is no sample rejection, and as a result there is likely to be less wasted computation. 
Qualitatively, the equivalent discrete time process $\mb{\hat{T}}$ can be expected to visit low probability states far more frequently than an unweighted sampler. Those states will just have very short holding times, and be assigned very small importance weights. This will allow it to more rapidly explore the state space.

\subsection{Hamiltonian Monte Carlo}

\begin{figure}
\hspace{-5mm}
%\parbox[l]{0.25\linewidth}{
\begin{tabular}[b]{cc}
\begin{tabular}[b]{l}%
	\includegraphics[width=0.45\linewidth]{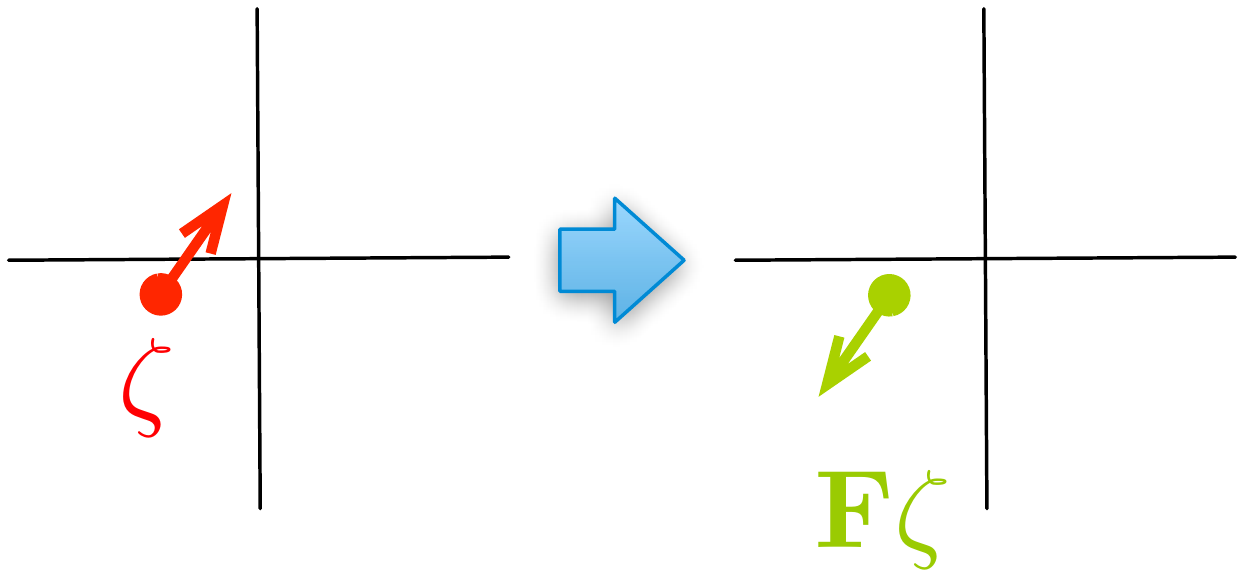}\\
	% scale this one by 0.475/0.45, because L\zeta sticks out
	\includegraphics[width=0.475\linewidth]{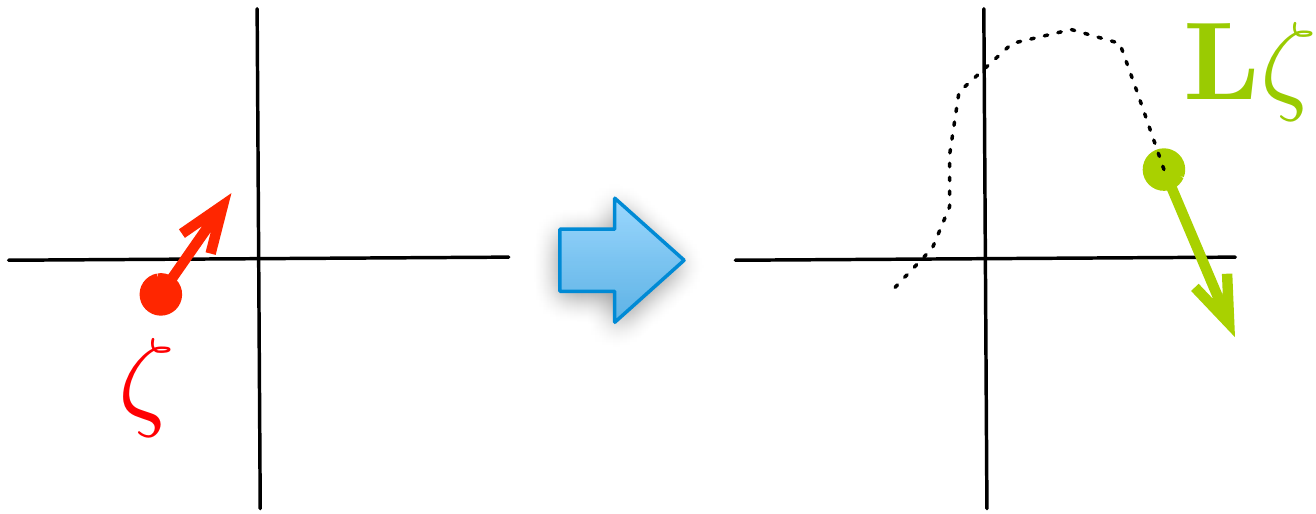}\\
	\includegraphics[width=0.45\linewidth]{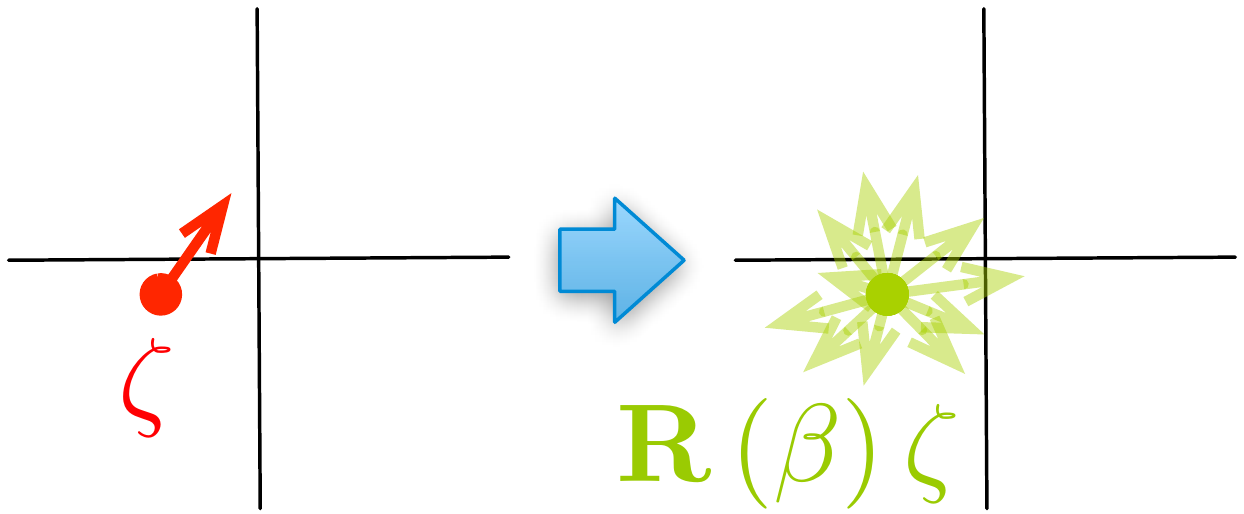}\\(a)
\end{tabular}
&
\begin{tabular}[b]{l}
% NOTE this image appears in the wrong place in sharelatex sometimes
% it's in the right place when you download the PDF though, so don't
% worry about it
	\includegraphics[width=0.45\linewidth]{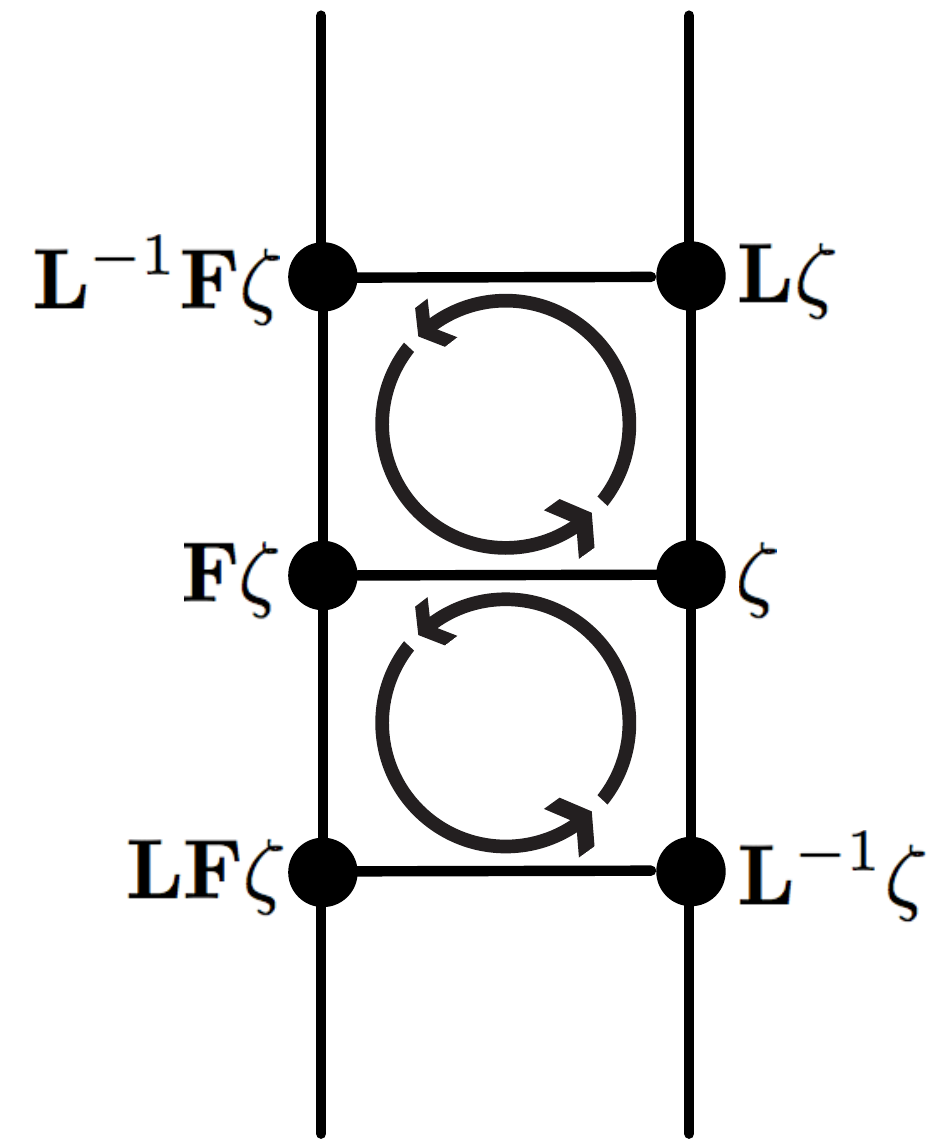}
	\\(b)
\end{tabular}
\end{tabular}
%} 
    \caption{
{\em (a)} The action of operators involved in Hamiltonian Monte Carlo (HMC).  The base of each {red or green arrow} represents the position $\mb x$, and the length and direction of {each of these arrows} represents the momentum $\mb v$.  The flip operator $\mb F$ reverses the momentum.  The leapfrog operator $\mb L$ approximately integrates Hamiltonian dynamics.  The trajectory taken by $\mb L$ is indicated by the dotted line.  
The randomization operator $\mb R$ replaces the momentum.
{\em (b)}    
    The ladder of discrete states generated by the leapfrog ($\mb L$) and flip ($\mb F$) operators. 
        Application of $\mb F$ corresponds to movement across the `rungs' of the ladder. Application
        of $\mb L$ corresponds to movement up the right side of the ladder, or down the left. 
        Inset arrows illustrate closed loops of constant total probability flow under our chosen rate 
        (Equation~\ref{eq loop condition})
        }
    \label{fig ladder}
\end{figure}

Hamiltonian Monte Carlo (HMC) \cite{Duane1987,Neal:HMC} is the state-of-the-art, general purpose Monte Carlo algorithm for sampling from a distribution $\pi\left(\mb x\right)$ over a continuous state space $\mb x \in \mathbb R^N$.

HMC utilizes the same Hamiltonian dynamics that govern the evolution of a physical system -- for instance a marble rolling in a swimming pool -- to rapidly traverse long distances in state space.
In HMC the state space is first extended to include auxiliary momentum variables $\mb v \in \mathbb R^N$ with distribution $\pi\left(\mb v\right)$, such that the joint state space over position and momentum is $\zeta = \{\mb x, \mb v\}$, with joint distribution $\pi\left(\zeta\right) = \pi\left(\mb x\right)\pi\left(\mb v\right)$.
An analogy is then made between $\mb x$ and physical position (e.g. the position of the marble), between $\mb v$ and physical momentum (the momentum of the marble), and between $(-\log\pi\left(\mb x\right))$ and potential energy (the height of the swimming pool at position $\mb x$).
Since physical dynamics conserve energy,
they can generate very long trajectories in state space while remaining on a constant probability contour of $\pi\left(\mb zeta\right)$.

HMC is thus able to move very long distances in state space in a single update step.

\subsubsection{Discrete state space Ladder} \label{sec state ladder}

As introduced in \cite{sohl2014hamiltonian} and illustrated in Figure \ref{fig ladder}, HMC can be viewed in terms of transitions on a discrete state space ladder.
This state ladder is formulated by expressing the action of HMC on a sampling particle in terms of three operators.
The leapfrog integration operator, $\mb L$, approximately integrates Hamiltonian dynamics for a fixed number of time steps and a fixed step length.

The momentum flip operator, $\mb F$, reverses a particle's direction of travel along a contour by flipping its momentum.
The momentum randomization operator $\mb R$ redraws the momentum vector from $\pi\left(\mb v\right)$, and moves a particle onto a new state space ladder.

This perspective suggests a powerful formalism, effectively discretizing the state space, and it illuminates the structure\footnote{
  Between momentum randomizations, HMC acts in a manner isomorphic to the Dihedral group of, in general, infinite order.
  The HMC state ladder, and thus the order, is generally infinite because trajectories through state space produced by Hamiltonian dynamics are almost never closed.
 
}
of HMC.
Throughout this work, we refer to the structure generated by the operators as a state ladder.
As illustrated in Figure \ref{fig ladder}, $\mb L$ causes movement up the right side of a state ladder
and down the left side,
whereas $\mb F$  causes horizontal movement across the rungs of a ladder. A trajectory can be exactly reversed by reversing the momentum, integrating Hamiltonian dynamics, and reversing the momentum again. As can be seen in Figure \ref{fig ladder}, this corresponds to making a loop on the state space ladder, and it implies that $\mb F \mb L \mb F \mb L = \mb I$, where $\mb I$ is the identity operator.
$\mb R$ causes movement off of the current state ladder and onto a new one.
Both $\mb F$ and $\mb L$ are volume preserving, which will eliminate the need to consider volume changes when computing Markov transition rates.

If we only allow transitions between states that are connected on the state ladder (Figure \ref{fig ladder}), then transitions can only occur between $\mb \zeta$ and three other states ($\mb L^{-1}\zeta$, $\mb F\zeta$, $\mb L\zeta$). %$\mb R\zeta$ will be discussed separately. 
This makes HMC well matched to Markov jump processes, since only a small number of transitions need be considered.

\subsubsection{Current research} \label{sec current research}
The development of improved methods for performing HMC is an important area of active research. These include the use of shadow Hamiltonians that are more closely conserved by the approximate Hamiltonian integrator \cite{Izaguirre2004},
Riemann manifold HMC \cite{Girolami2011a} and other investigation of its geometry \cite{betancourt2014geometric}, quasi-Newton HMC \cite{Zhang2011}, Hilbert space HMC \cite{Beskos2011}, parameter adaptation techniques \cite{Wang2013}, Hamiltonian annealed importance sampling \cite{HAIS}, split HMC \cite{Shahbaba2011}, tempered trajectories \cite{Neal:HMC}, novel discrete time transition rules \cite{sohl2014hamiltonian,Sohl-Dickstein2012,campos2015extra}, stochastic gradient variants on HMC \cite{chen2014stochastic}, HMC for approximate Bayesian computation \cite{meeds2015hamiltonian}, and new approximate Hamiltonian integrators \cite{chaoexponential}.

\section{Markov jump Hamiltonian Monte Carlo}

\begin{figure*}
    \centering      
    \includegraphics[width=0.7\linewidth]{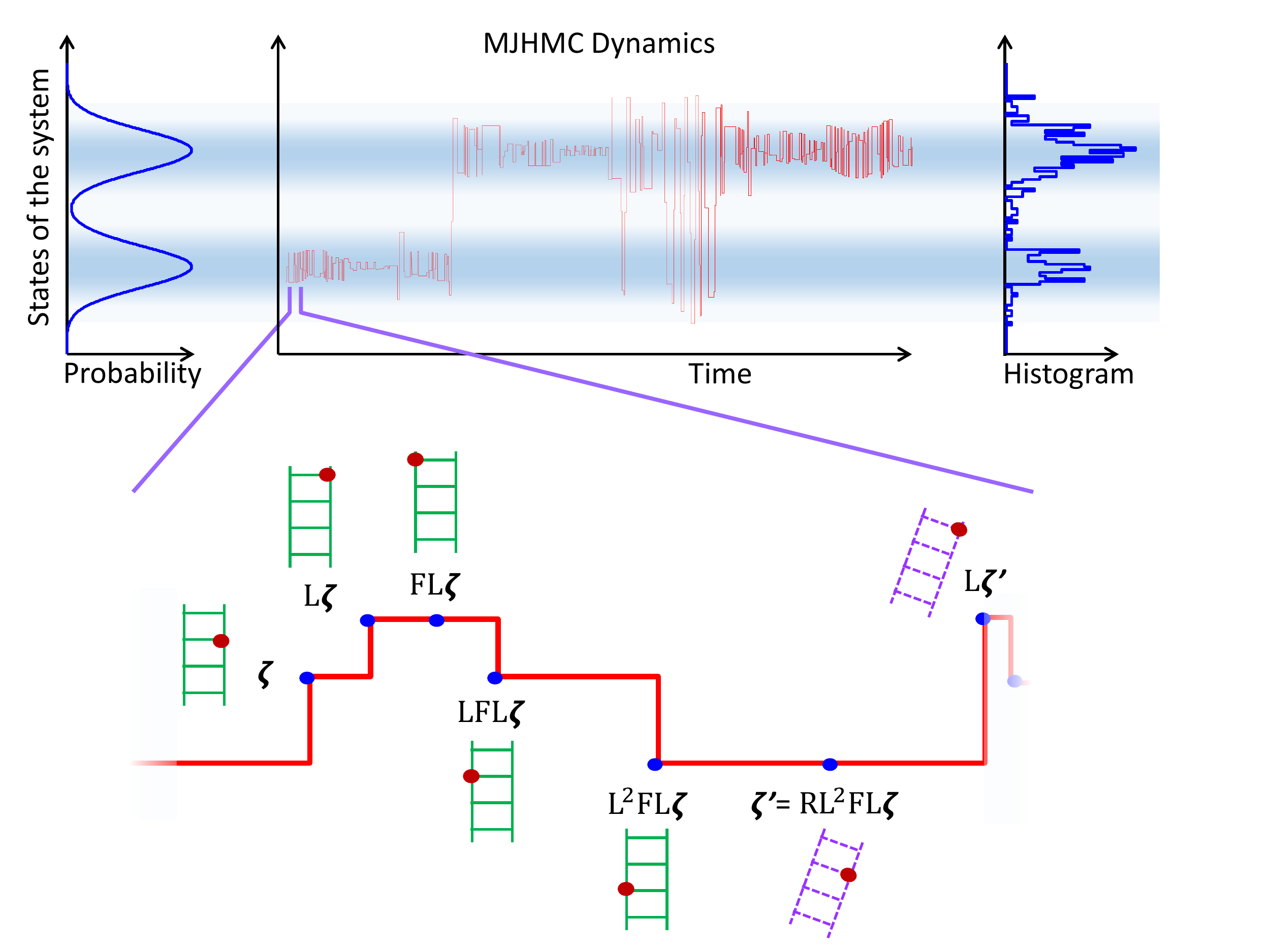}
    %\title{Illustration of Markov Jump HMC sampling dynamics generated from real sampling data.}
    \caption{Illustration of Markov Jump HMC sampling dynamics. 
    %generated from real sampling data. 
        The red curve represents a particle trajectory for 400 time steps. Blue shading indicates the probability density, plotted at the left, and an empirical histogram of the samples is shown at the right.  The inset blowup at the bottom shows how movement of the sampling particle corresponds to transitions on the state ladder, using the symbolic and graphical conventions described in Figure~\ref{fig ladder}. Note that the sampling particle dwells in a position for a duration related to the probability density of that state relative to its neighbors. We have indicated the transition from one state ladder (vertical green ladder) to a new ladder (angled purple ladder) following a momentum randomization event, resulting in a new state labeled $\zeta^\prime$. 
}
    \label{fig:MJHMC_illustration}
\end{figure*}

We now discuss various aspects of our MJHMC sampler, 
schemetized in Figure~\ref{fig:MJHMC_illustration}. 
\subsection{Continuous time transition rates on HMC state ladder}

A Markov process must satisfy \emph{two conditions} \cite{Karlin1968} to sample from a target distribution .
The first is ergodicity, which requires that the process will eventually explore the full state
space; this is typically straightforward to satisfy.

The second condition is that the target distribution must be a fixed point of the Markov process. This is usually achieved via dynamics that satisfy detailed balance matched to the state probabilities of the target distribution.

\subsubsection{Closed loops preserve the fixed point distribution}

Markov transition rates $\mb{\tilde{\Gamma}}\left(\zeta' \mid \zeta\right)$ that preserve $\pi\left(\zeta\right)$ as a fixed point can be constructed from
closed loops of constant probability flow in state space.

For 
closed loops to have constant probability flow, the flow $r\left(\zeta_i \mid \zeta_j \right) = \Gamma\left(\zeta_i \mid \zeta_j \right)\pi\left( \zeta_j \right)$ from state $j$ to $i$ must be identical for each link % $\{i,j\}$
in the 
loop.
This is analogous to Kirchoff's current law for electrical circuits -- constant probability flow in all loops implies that the net probability flow into any state equals the net 
flow out of that state.

\subsubsection{Choosing transition rates}
\label{sec loop rates}

For the case of HMC, we set the loops to be between the ``rungs" of the state ladder, as illustrated in Figure \ref{fig ladder}. The loop balance condition for each closed loop becomes
\begin{align}
  \tilde{r}(\mb F \zeta \mid \zeta) = \tilde{r}(\mb{LF} \zeta \mid \mb F\zeta) = \tilde{r}(\mb{L^{-1}} \zeta \mid \mb{LF} \zeta) = \tilde{r}(\zeta \mid \mb L^{-1} \zeta),
\end{align}
\begin{align}
  \label{eq loop condition}
  \pi\left(\zeta\right) \tilde{\Gamma}\left(\mb F \zeta \mid \zeta\right) &
  = \pi\left(\mb F \zeta\right) \tilde{\Gamma}\left(\mb{LF} \zeta \mid \mb F\zeta\right) \nonumber \\ &
  = \pi\left(\mb{LF} \zeta\right) \tilde{\Gamma}\left(\mb{L^{-1}} \zeta \mid \mb{LF} \zeta\right) \nonumber \\ &
  = \pi\left(\mb{L^{-1}} \zeta\right) \tilde{\Gamma}\left(\zeta \mid \mb L^{-1} \zeta\right)
  .
\end{align}

In order to satisfy this condition, we set the transition rates to be
\begin{align}
\label{eq MH}
\tilde{\Gamma}(\zeta' \mid \zeta) &=
	\left\{\begin{array}{ccc}
		\left[\frac{\pi\left(\mb L \zeta\right)}{\pi\left(\zeta\right)}\right]^{\frac{1}{2}} & & \zeta' = \mb L \zeta \\
		\left[\frac{\pi\left(\mb L \mb F \zeta\right)}{\pi\left(\mb F\zeta\right)}\right]^{\frac{1}{2}} & & \zeta' = \mb F \zeta \\
		0 & & \text{otherwise}
	\end{array}\right.
.
\end{align}
One can verify by direct substitution that these transition rates satisfy Equation \ref{eq loop condition}.
The transition rates for the full ladder consist of a sum over the transition rates for each loop.

\subsubsection{Opposing flows cancel}
\label{sec opposing}

As can be seen in Figure \ref{fig ladder}, adjacent loops make flip transitions across the ``rungs" of the ladder in opposite directions. After summing over all loops, the net transitions across the ladder approximately cancel.

This allows us to reduce the flip rates in both directions, such that the flip rate in one direction is zero. The final rate of flip transitions in our algorithm will thus be

\begin{align}
  \label{eq reduced rate}
  \Gamma(\mb F\zeta \mid \zeta) &= \tilde{\Gamma}(\mb F \zeta \mid \zeta) - \min\bracket{\tilde{\Gamma}(\mb F \zeta \mid \zeta), \tilde{\Gamma}(\zeta \mid \mb  F \zeta)},
  \\
   &= \max\bracket{ 0,
  \tilde{\Gamma}(\mb F \zeta \mid \zeta) - \tilde{\Gamma}(\zeta \mid \mb  F \zeta)},
  \\
   &= \max\bracket{ 0,
        \left[\frac{\pi\left(\mb L \mb F \zeta\right)}{\pi\left(\mb F\zeta\right)}\right]^{\frac{1}{2}}
            - \left[\frac{\pi\left(\mb L \mb F \mb F \zeta\right)}{\pi\left(\mb F\mb F\zeta\right)}\right]^{\frac{1}{2}}
        },
  \\
  \label{eq reduced rate final}
   &= \max\bracket{ 0,
        \left[\frac{\pi\left(\mb L^{-1}\zeta\right)}{\pi\left(\zeta\right)}\right]^{\frac{1}{2}}
            - \left[\frac{\pi\left(\mb L \zeta\right)}{\pi\left(\zeta\right)}\right]^{\frac{1}{2}}
        },
\end{align}

where the final step relies on the observations that $\mb F \mb F \zeta = \zeta$, and that  $\pi\left(\zeta\right) = \pi\left(\mb F\zeta\right)$ \cite{sohl2014hamiltonian}. In practice, $\pi\left(\mb L^{-1}\zeta\right)$ will typically already be available (up to a shared normalization constant) from the preceding Markov transition, and will not need to be computed.

Due to discretization error, the leapfrog integrator for Hamiltonian dynamics only approximately conserves probability.
Equation \ref{eq reduced rate final} shows that the residual flow across the ladder stems from this discretization error of the leapfrog integrator.
This is completely analogous to the cause of momentum flips in standard HMC.

\subsection{Momentum randomization}
\label{sec momentum}

In discrete time HMC the momentum is periodically corrupted with noise.
If this was not done, then sampling would be restricted to a single state ladder, and mixing between state ladders would not occur.
In order to accomplish the same end in continuous time, we jump to a state $\mb R \zeta$ with a constant transition rate $\beta$. A transition to $\mb R \zeta$ corresponds to replacing the momentum $\mb v$ with a new draw from $\pi\left(\mb v\right)$.
The transition rate from $\mb v$ to a particular $\mb v'$ is thus $\beta \pi\left(\mb v'\right)$.
It can be seen by substitution that this rate satisfies detailed balance. 

\subsection{Final transition rates}

Combining the transition rates derived in Sections \ref{sec loop rates}, \ref{sec opposing}, and \ref{sec momentum},
\begin{align}
\label{eq_MJHMC}
\mkern-48mu % negative horizontal space so full equation can be seen. ?? TODO may need to change this as text added/removed
\Gamma(\zeta' \mid \zeta) &=
	\left\{\begin{array}{ccc}
		\left[\frac{\pi\left(\mb L \zeta\right)}{\pi\left(\zeta\right)}\right]^{\frac{1}{2}} & & \zeta' = \mb L \zeta \\
		 \max\bracket{ 0,
        \left[\frac{\pi\left(\mb L^{-1}\zeta\right)}{\pi\left(\zeta\right)}\right]^{\frac{1}{2}}
            - \left[\frac{\pi\left(\mb L \zeta\right)}{\pi\left(\zeta\right)}\right]^{\frac{1}{2}}
        } & & \zeta' = \mb F \zeta \\
		\beta & & \zeta' = \mb R \zeta \\
		0 & & \text{otherwise}
	\end{array}\right.
.
\end{align}
We verify that these transition rates satisfy the balance condition for $\pi\left(\zeta\right)$ in Appendix \ref{app balance}. 
The third line is a slight abuse of notation since $\mb R \zeta$ does not correspond to a single fixed state, but rather indicates that the momentum is replaced by a new draw from $\mb \pi\left(\mb v\right)$, where this replacement is triggered by a Poisson process with rate $\beta$. 
Note that as in \cite{sohl2014hamiltonian} these dynamics do not satisfy detailed balance, and can be expected to mix more quickly as a result \cite{Ichiki2013}.

\subsection{System time vs compute time}\label{sec sys comp}

We have described continuous time dynamics in terms of a system, or simulation, time.
However, when applying this sampler to a real problem it is its performance as measured relative to compute time that matters.
Here we show how to relate the continuous time dynamics of the Markov jump process to a discrete time Markov process, with an approximately fixed computational cost per time-step.

First we observe that there is a discrete time Markov process describing only the sequence of visited states, thus neglecting the holding time spent in each state. For notational convenience we represent Markov processes using matrix notation in this section. The update rule for this Markov process can be written
\begin{align}
\label{eq pbar evolve}
\mb{\hat{p}}^{\tau+1} &= \mb{\hat{T}} \mb{\hat{p}}^\tau, \\
\label{eq pbar relation}
\hat{T}_{ij} &= 
    \left\{\begin{matrix}
    \prod_{k \neq j} \frac{\Gamma_{ij}}{\Gamma{ij} + \Gamma{ik}}
    %     \prod_{k \neq j}\frac{\Gamma_{ij}}{\Gamma{ij} + \Gamma{ik}}
            & & i \neq j \\
        0
            & & i = j
	\end{matrix}\right.
\end{align}

where the matrix $\mb{\hat{T}}$ is the Markov transition kernel, and the vector $\mb{\hat{p}}^\tau$
is the probability distribution over system states at timestep $\tau$. The computational cost of each time step is roughly constant under this Markov chain, since each step requires computing the transition rate to all possible next states in Equation \ref{eq MH}.

The current and 
%equilibrium 
fixed point distributions $\mb{\hat{p}}$ and $\mb{\hat{\pi}}$ under this process can be related to the corresponding distributions  $\mb{p}$ and  $\mb{\pi}$ under the Markov jump process by scaling by the expected holding time,
\begin{align}
\label{eq_scale}
\mb{\pi} &= \frac{1}{Z}\mb D \mb{\hat{\pi}}, \\
\mb{p} &= \frac{1}{Z}\mb D \mb{\hat{p}},
\end{align}
where $\mb D$ is a diagonal matrix with the expected holding times for each state on the diagonal, $D_{jj} = \frac{1}{\sum_{i\neq j} \Gamma_{ij}}$, and $Z$ is a normalization constant\cite{Balakrishnan1996}.
Similarly, the evolution of $\mb p$ relative to these discrete time steps can be expressed by scaling $\mb{\hat{p}}$ in Equation \ref{eq pbar evolve} by the holding time,
\begin{align}
\mb{p}^{\tau+1} &= \mb D \mb{\hat{T}} \mb D^{-1} \mb{p}^\tau, \label{eq trans sim} \\
\mb{p}^{\tau+1} &= \mb T \mb{p}^\tau,
\end{align}
where $\mb T = \mb D \mb{\hat{T}} \mb D^{-1}$ describes the discrete time evolution of the samples. 
Since $\mb T$ and $\mb{\hat{T}}$ are related to each other by a similarity transform, they share identical eigenvalues. 
In order to evaluate the spectral gap, and thus the mixing time, of the Markov jump process in terms of computational time, it is thus sufficient to compute the spectral gap of $\mb{\hat{T}}$. 
We do this for randomly generated toy systems in Section \ref{sec spec gap} and Figure \ref{fig_spec_gap} and show that MJHMC has superior spectral gap characteristics, indicating more efficient mixing.

\subsection{Algorithm}

Here, we briefly summarize the Markov Jump HMC algorithm for generating $N$ samples in pseudocode. As with all HMC sampling algorithms, an energy function $E(x)$ (equivalent to $-\log \pi\left(\mb x\right)$ plus a constant) and its gradient are required. The three hyperparameters are the leapfrog step size $\epsilon$, the number of leapfrog steps per sampling step $M$; and the momentum corruption rate $\beta$.

Note that computation of $\mb L^{-1} \zeta$ is only necessary when the last transition made was a momentum flip or randomization. The number of times the gradient is evaluated in an MJHMC sampling step is comparable to that of standard HMC. 

\RestyleAlgo{boxruled}
\LinesNumbered
\IncMargin{2em}
\begin{algorithm}[]
\caption{Markov Jump Process HMC}
\SetKwInOut{Input}{input}
\SetKwInOut{Output}{output}

\Input{$\epsilon$,$M$,$\beta$,$E(x)$,$\nabla E(x)$, $N$}
\Output{$N$ samples}
$\zeta_0 \leftarrow \text{Randomly initialized}$ \;
%$h_0 \leftarrow \text{Initialize holding time}$ \;
\For{$i \leftarrow 1$ \KwTo $N$}{
 Calculate states $\mb L \zeta_{i-1}, \mb F \zeta_{i-1}, \mb L^{-1} \zeta_{i-1}$ \;
 Compute $E\paren{\zeta_{i-1}}$, $E\paren{\mb L \zeta_{i-1}}$, $E\paren{\mb F \zeta_{i-1}}$, $E\paren{\mb L^{-1} \zeta_{i-1}}$ \;
 Compute transition rates $\Gamma_L$, $\Gamma_{F}$, $\Gamma_R$  using Equation \eqref{eq_MJHMC} \;
 Draw waiting times $w_L$, $w_F$, $w_R$ from an exponential distribution, using rates of $\Gamma_L$, $\Gamma_{F}$, $\Gamma_R$ respectively\;
 Record holding time for $\zeta_{i-1}$. $h_{i-1} \leftarrow \min(w_L, w_F, w_R)$ \;
 Set $\zeta_i$ to whichever of $\mb L\zeta$, $\mb F\zeta$, $\mb R\zeta$ had the shortest waiting time \;
 } 
 Resample all $\zeta_i$ using holding times $h_i$ as importance weights \;
 \end{algorithm}

\section{Experimental results} \label{sec exp}

\subsection{Spectral gap on HMC state ladder}\label{sec spec gap}

The convergence rate of a Markov process to its steady state is given by its spectral gap\cite{Wilmer}. This is the difference in the magnitude of the two largest eigenvalues. 
We numerically compute this value for randomly generated toy problems in order to compare our mixing rate to that of standard HMC.
As all HMC algorithms randomize momentum in nearly the same way, it is expected that their mixing time over a single state ladder is representative of their mixing time over the entire state space. To achieve analytic tractability we restrict our attention to finite state ladders.
To avoid edge effects, we attach the top and bottom rungs of the ladder to each other, so that the ladder forms a loop and $\mb L^k \zeta = \zeta$, where $k$ is the number of distinct rungs. 
We evaluate the eigenvalues on each state ladder using the similarity relationship to a discrete time Markov chain in Equation \ref{eq trans sim}.

\begin{figure}[t]
  \centering
  \includegraphics[width=0.5\textwidth]{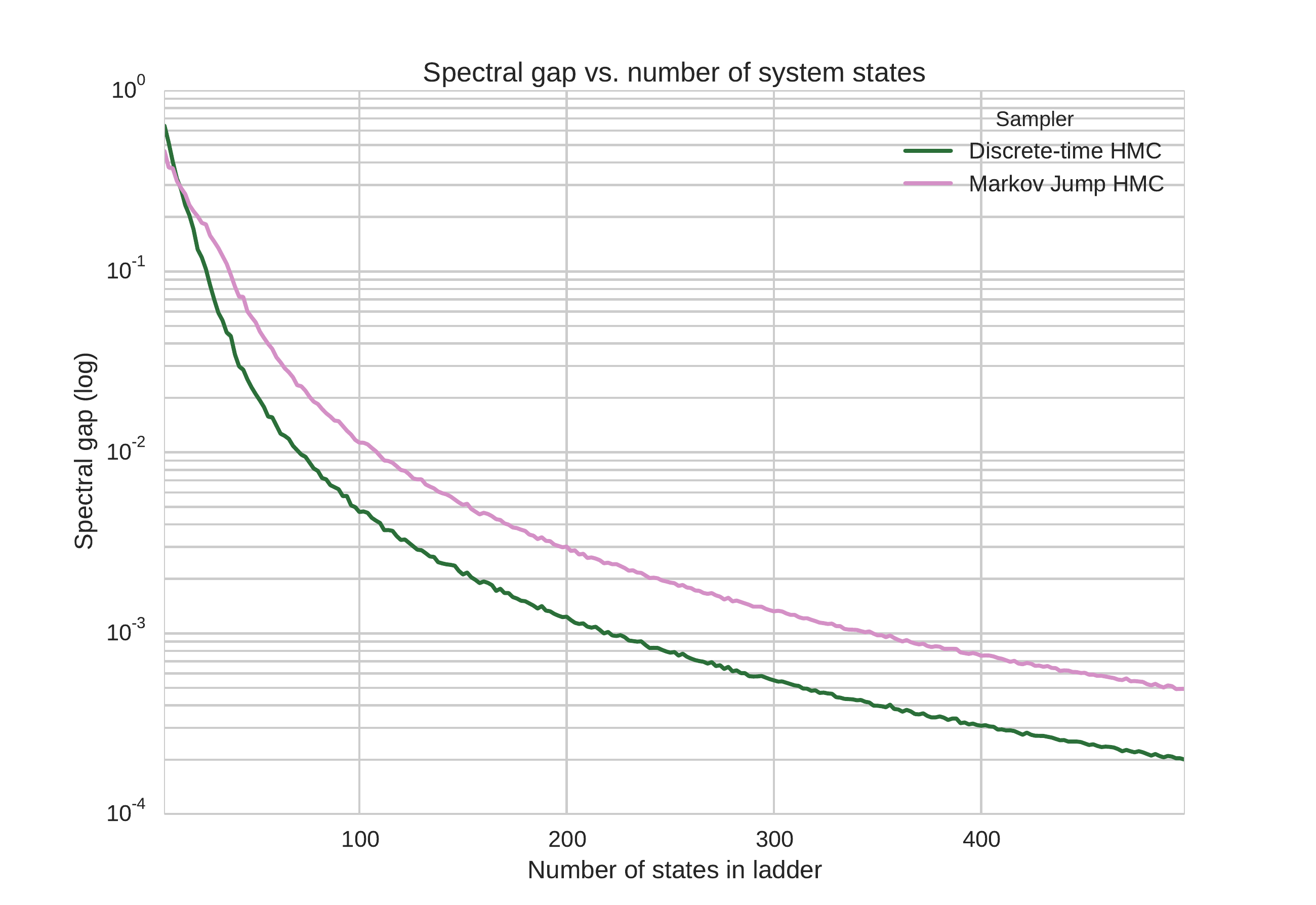}
  \caption{Comparison of mixing performance of Markov jump HMC (MJHMC) and standard discrete time HMC. Spectral gap versus size of state ladder. For large state ladder sizes, MJHMC is better by half an order of magnitude.\label{fig_spec_gap}}
  \end{figure}

A comparison of such spectral gaps between Markov Jump HMC 
and standard HMC is illustrated in Figure \ref{fig_spec_gap} as a function of state ladder size. 
%Rather than use the energy function of a physical system, 
We draw the energy for each `rung' of the state space ladder from a unit norm Gaussian distribution, and average across 250 draws for each ladder size.
Figure \ref{fig_spec_gap} thus shows performance averaged over many randomly generated energy landscapes.
MJHMC mixes faster (has a larger spectral gap) for all except the smallest state space sizes.

  \begin{figure}[t]
  \centering
  \includegraphics[width=0.5\textwidth]{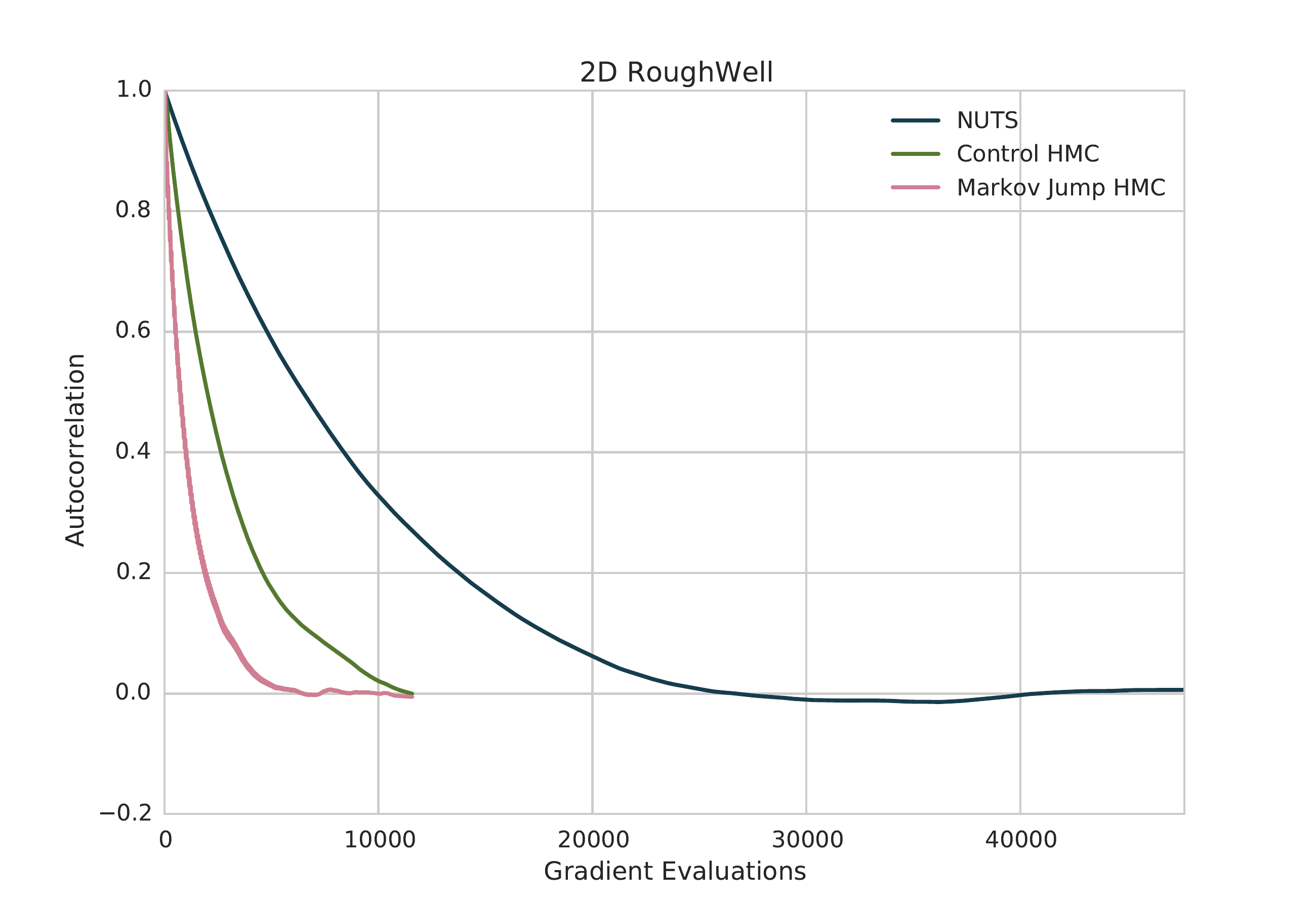}
    \caption{Autocorrelation versus number of gradient evaluations for standard HMC and MJHMC for the Rough Well distribution. The hyperparameters found by  Spearmint for MJHMC are $\eps = 3.0, \beta = 0.012314, M = 25$ and for control HMC $\eps = 0.591686, \beta = 0.429956, M=25$.\label{fig_ac}}
  \end{figure}

\subsection{Autocorrelation on rough well distribution}

Explicit computation of mixing time for most problems is computationally intractable. 
It is common to instead use the rate at which the sample autocorrelation approaches zero as a proxy. 
As illustrated in Figure \ref{fig_ac}, we compare autocorrelation traces for MJHMC with standard HMC and NUTS on the rough well distribution, and find that MJHMC performs significantly better.

Our results, illustrated in Figure \ref{fig_ac}, indicate that MJHMC significantly outperforms standard HMC. % when each sampler is set to the best hyperparameters found with Spearmint. 

\subsubsection{Energy function}

The chosen energy function was
the `rough well' distribution from \cite{sohl2014hamiltonian}.
This distribution provides a good test case because it is as simple as possible, while also presenting both well understood and significant challenges to HMC-style samplers.
Its energy function is
\begin{align}
E(x) &= \frac{1}{2 \sigma_1^2}(x_{1}^{2} + x_{2}^{2}) + cos\left(\frac{\pi x_1}{\sigma_2}\right) + cos\left(\frac{\pi x_2}{\sigma_2}\right),
\end{align} 
where $\sigma_1 = 100 $ and $\sigma_2 = 4$. Although this distribution is well conditioned everywhere, the sinusoids cause it to have a `rough' surface, such that it requires many leapfrog steps to traverse the quadratic well while maintaining a reasonable discretization error.

\subsubsection{Hyperparameter selection}
\label{sec spearmint}
As HMC samplers are sensitive to the choice of hyperparameters, we searched for
optimal settings of the hyperparameters for all samplers with the Spearmint package \cite{snoek2012practical}.

For each hyperparameter setting, we computed the autocorrelation as a function of number of gradient evaluations $n$, $C\left( n \right)$. We then fit a function $\rho\left(n; r\right)$ to this in terms of a complex coefficient $r \in \mathbb{C}$, 
\begin{align}
\hat{r} &= \argmin_r \left|\left| \rho\left(n; r\right) - C\left( n\right)\right|\right|^2 \\
\rho\left(n; r\right) &= \operatorname{Re}\left[\exp\left( r n \right)\right]
\end{align} 
where the imaginary portion of $r$ corresponds to an oscillatory rate, and the real part corresponds to the decay rate towards 0 autocorrelation. We use $\operatorname{Re}\left(r\right)$ as the Spearmint objective.

We provide additional figures to provide support for our hyperparameter search results in Appendix Figures  \ref{fig search contr}, \ref{fig search beta}, and \ref{fig search M}. 

NUTS self-tunes its hyperparamters during burn-in, so we did not perform a hyperparameter search for NUTS.

% \subsection{Mixing in Product of Experts distribution}

%% Hey you, reading the latex source on arxiv! This experiment is coming. We ran out of time to run it in a convincing way (with Spearmint hyperparameter optimization) before the AISTATS deadline.

% In order to test performance on a higher dimensional energy landscape, we evaluated mixing on a Product of Experts (PoE) distribution, with Student's t-experts \cite{welling2002learning}. As is illustrated in Figure \ref{fig poe}, MJHMC mixed more quickly when sampling from the PoE model.

% The PoE distribution was over $6\times 6$ image patches, with 72 experts. The energy function was,
% \begin{equation}
% E(x) = \sum_{i} \alpha_{i} log(1+(W_{i}^{T}X)^2)
% \end{equation}

% where the receptive fields $\mb W$ were set such that the Energy function had zero mean from a random uniform distribution such that only 25$\%$ of the entries were active. $\alpha$ was chosen from a random normal distribution. 

% Hyperparameters were chosen using Spearmint, as described in Section \ref{sec spearmint}.

%   \begin{figure}[t]
%   \centering
%   \includegraphics[width=0.5\textwidth]{figures/}
%     \caption{
%     Mixing for a Product of Experts distribution with Student's t-experts over $6\times 6$ pixel image patches. The top row corresponds to MJHMC, and the bottom row corresponds to standard HMC. Each column corresponds to the same sampling chain after a different number of gradient evaluations. Note that the MJHMC samples transform more quickly, corresponding to more rapid mixing.
%     \label{fig poe}}
%   \end{figure}

\section{Discussion}

We introduced an algorithm, Markov Jump Hamiltonian Monte Carlo (MJHMC), in which the state transitions in Hamiltonian Monte Carlo sampling occur as Poisson processes in continuous time, rather than at discrete time steps. We demonstrated that this algorithm led to improved mixing performance, as measured by explicit computation of the spectral gap, by the autocorrelation of the sampler on a simple but challenging distribution.%, and by illustration of sequences of samples on a high dimensional product of experts model.

\section{Acknowledgements}
MRD and MM were both supported by NSF grant IIS-1219199 to Michael R. DeWeese. MM was also partially supported by NGA grant HM1582-081-0007 to Bruno Olshausen and NSF grant IIS-1111765 to Bruno Olshausen.   ABB was partially supported by a Berkeley Physics Undergraduate Research Scholars Program Fellowship.  This material is based upon work supported in part by the U.S. Army Research Laboratory and the U.S. Army Research Office under contract number W911NF- 13-1-0390. 

\small

\bibliographystyle{unsrt}
% \bibliography{jascha_library,andrew_library}
\bibliography{continuous_hmc}

\newpage
\onecolumn
\normalsize
\appendix
\part*{Appendix}

\setcounter{figure}{0} \renewcommand{\thefigure}{A.\arabic{figure}}
\setcounter{table}{0} \renewcommand{\thetable}{A.\arabic{table}}

\section{Transition rates satisfy balance condition}
\label{app balance}

The continuous time balance condition states that, at the steady state distribution, there is no net change in the probability of states,
\begin{align}
\pd{
    p\left( \zeta \right)
    }{t}
\Bigg|_{
    p\left( \zeta \right) = \pi\left( \zeta \right)
    }
= 0
.
\end{align}
In order to demonstrate that we satisfy the balance condition, we evaluate $\pd{
    p\left( \zeta \right)
    }{t}
\Bigg|_{
    p\left( \zeta \right) = \pi\left( \zeta \right)
    }$ using the transition rates from Equation \ref{eq_MJHMC},
\begin{align}
\pd{
    p\left( \zeta \right)
    }{t}
\Bigg|_{
    p\left( \zeta \right) = \pi\left( \zeta \right)
    }
= & 
-\pi\left( \zeta \right) \left[\frac{\pi\left(\mb L \zeta\right)}{\pi\left(\zeta\right)}\right]^{\frac{1}{2}}
\nonumber \\ &
+\pi\left( \mb L^{-1} \zeta \right) \left[\frac{\pi\left(\mb L \mb L^{-1} \zeta\right)}{\pi\left(\mb L^{-1}\zeta\right)}\right]^{\frac{1}{2}}
\nonumber \\ &
- \pi\left( \zeta \right) \max\bracket{ 0,
        \left[\frac{\pi\left(\mb L^{-1}\zeta\right)}{\pi\left(\zeta\right)}\right]^{\frac{1}{2}}
            - \left[\frac{\pi\left(\mb L \zeta\right)}{\pi\left(\zeta\right)}\right]^{\frac{1}{2}}
        }
\nonumber \\ &
+ \pi\left( \mb F \zeta \right) \max\bracket{ 0,
        \left[\frac{\pi\left(\mb L^{-1}\mb F\zeta\right)}{\pi\left(\mb F\zeta\right)}\right]^{\frac{1}{2}}
            - \left[\frac{\pi\left(\mb L \mb F\zeta\right)}{\pi\left(\mb F\zeta\right)}\right]^{\frac{1}{2}}
        }
\nonumber \\ &
- \pi\left( \zeta \right) \beta
\nonumber \\ &
+ \int d\mb x' d \mb v' \pi\left(\mb x', \mb v' \right) \delta\left(\mb x'-\mb x\right) \pi\left( \mb v \right) \beta
,
\end{align}
where negative terms correspond to probability flow out of state $\zeta$ into other states, and positive terms correspond to probability flow from other states into state $\zeta$. There are only a small number of terms because transitions are only allowed to/from a limited set of states. We now proceed to simplify and cancel terms,
\begin{align}
\pd{
    p\left( \zeta \right)
    }{t}
\Bigg|_{
    p\left( \zeta \right) = \pi\left( \zeta \right)
    }
= & 
-\left[\pi\left(\mb L \zeta\right)\pi\left(\zeta\right)\right]^{\frac{1}{2}}
+\left[\pi\left(\zeta\right)\pi\left(\mb L^{-1}\zeta\right)\right]^{\frac{1}{2}}
\nonumber \\ &
- \pi\left( \zeta \right) \max\bracket{ 0,
        \left[\frac{\pi\left(\mb L^{-1}\zeta\right)}{\pi\left(\zeta\right)}\right]^{\frac{1}{2}}
            - \left[\frac{\pi\left(\mb L \zeta\right)}{\pi\left(\zeta\right)}\right]^{\frac{1}{2}}
        }
%\nonumber \\ &
+ \pi\left( \zeta \right) \max\bracket{ 0,
        \left[\frac{\pi\left(\mb L \zeta\right)}{\pi\left(\zeta\right)}\right]^{\frac{1}{2}}
            - \left[\frac{\pi\left(\mb L^{-1}\zeta\right)}{\pi\left(\zeta\right)}\right]^{\frac{1}{2}}
        }
\nonumber \\ &
- \pi\left( \zeta \right) \beta
+ \pi\left(\mb x \right) \pi\left( \mb v \right) \beta
\\
= & 
-\left[\pi\left(\mb L \zeta\right)\pi\left(\zeta\right)\right]^{\frac{1}{2}}
+\left[\pi\left(\zeta\right)\pi\left(\mb L^{-1}\zeta\right)\right]^{\frac{1}{2}}
\nonumber \\ &
+ \pi\left( \zeta \right) 
        \left[\frac{\pi\left(\mb L \zeta\right)}{\pi\left(\zeta\right)}\right]^{\frac{1}{2}}
            - \pi\left( \zeta \right)
        \left[\frac{\pi\left(\mb L^{-1}\zeta\right)}{\pi\left(\zeta\right)}\right]^{\frac{1}{2}}
\\
= & 
-\left[\pi\left(\mb L \zeta\right)\pi\left(\zeta\right)\right]^{\frac{1}{2}}
+\left[\pi\left(\zeta\right)\pi\left(\mb L^{-1}\zeta\right)\right]^{\frac{1}{2}}
+\left[\pi\left(\mb L \zeta\right)\pi\left(\zeta\right)\right]^{\frac{1}{2}}
-\left[\pi\left(\zeta\right)\pi\left(\mb L^{-1}\zeta\right)\right]^{\frac{1}{2}}
\\
= & 
0
.
\end{align}
Therefore the transition rates in MJHMC satisfy the balance condition for $\pi\left(\zeta\right)$, as claimed.

\section{Hyperparameter search}

\subsection{Demonstration of optimized hyperparameters}

  \begin{figure}[H]
  \centering
    \begin{subfigure}{.5\textwidth}
        \centering
        \includegraphics[width=\textwidth]{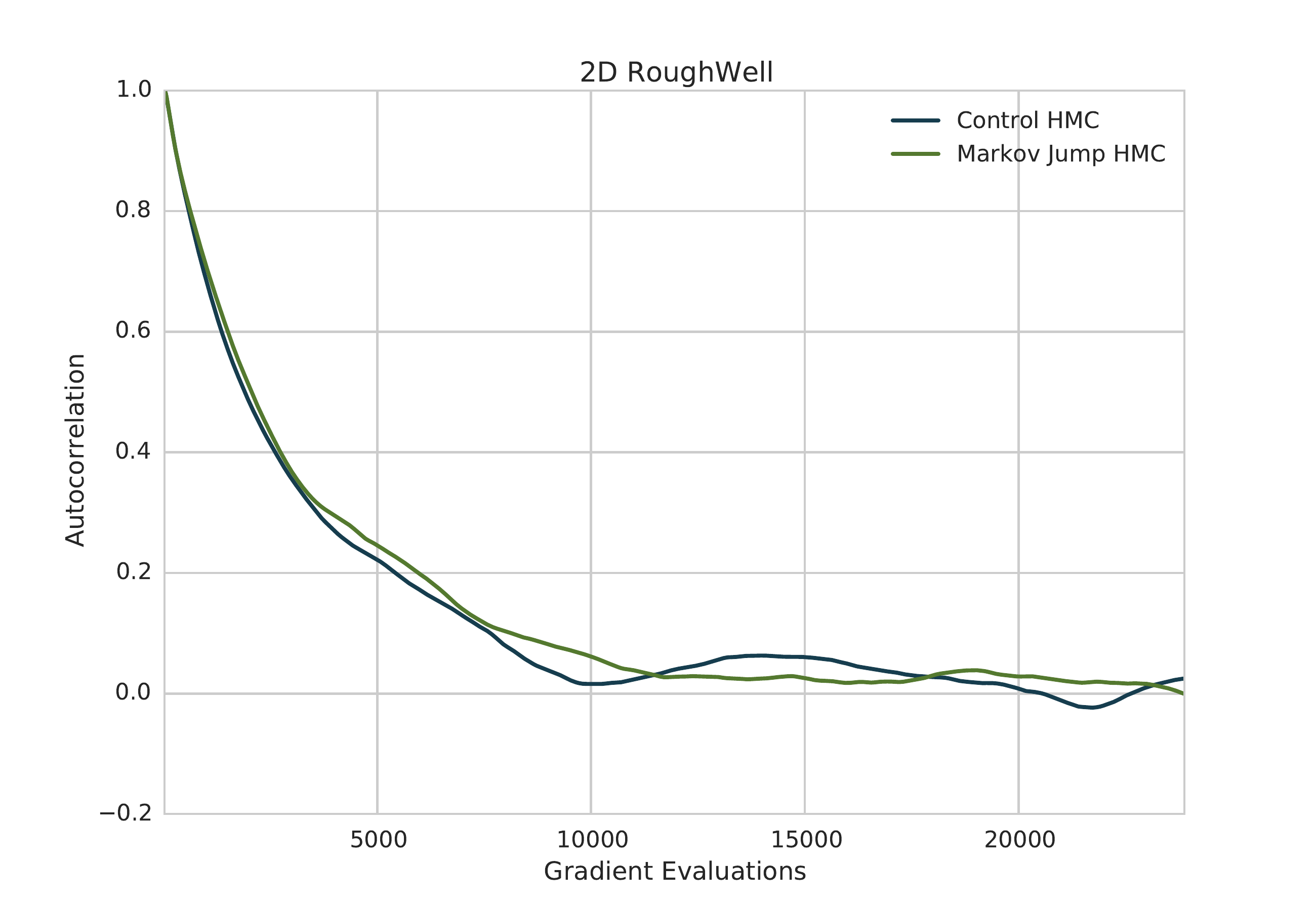}
        \label{fig search control}
    \end{subfigure}%
    \begin{subfigure}{.5\textwidth}
                \centering
        \includegraphics[width=\textwidth]{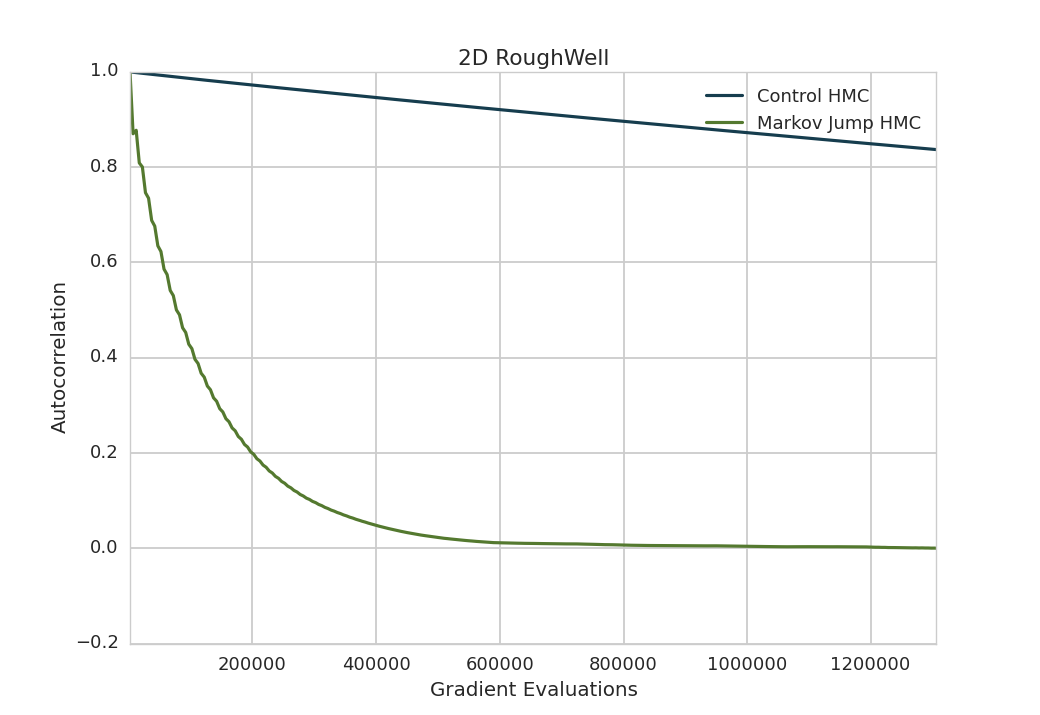}
        \label{fig search mjhmc}
    \end{subfigure}
  \caption{Comparison of mixing performance of MJHMC and standard discrete-time HMC
  with both samplers set to the same hyperparameters {\em (a)} Both samplers set to
    $\eps = 0.591686, \beta = 0.429956, M=25$, the best setting for standard HMC found by Spearmint
    {\em (b)} Both samplers set to $\eps = 3.0, \beta = 0.012314, M = 25$, the best settings for MJHMC found by Spearmint}
  \label{fig search contr}
  \end{figure}
  
    The autocorrelation data illustrated in figure  \ref{fig search contr} demonstrates that
    Spearmint found effective hyperparameters for MJHMC and standard discrete-time HMC on our 
    chosen energy function. Each sampler outperforms the other when both are set its 
    optimized setting of hyperparameters. 
    
\subsection{Illustration of Spearmint search}

\begin{figure}[H]
\centering
\includegraphics[width=.75\textwidth]{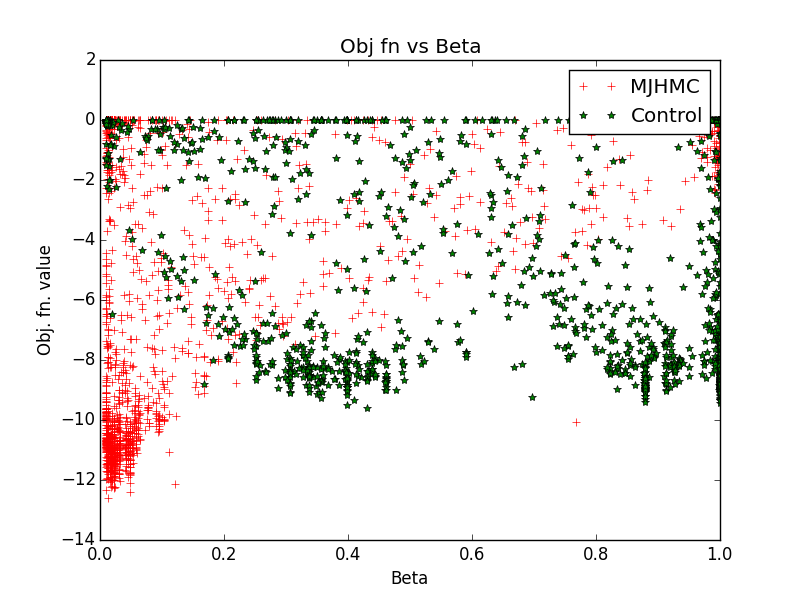}
\caption{Search performance projected onto the $\beta$ axis.
\label{fig search beta}}
\end{figure}

\begin{figure}[H]
\centering
\includegraphics[width=.75\textwidth]{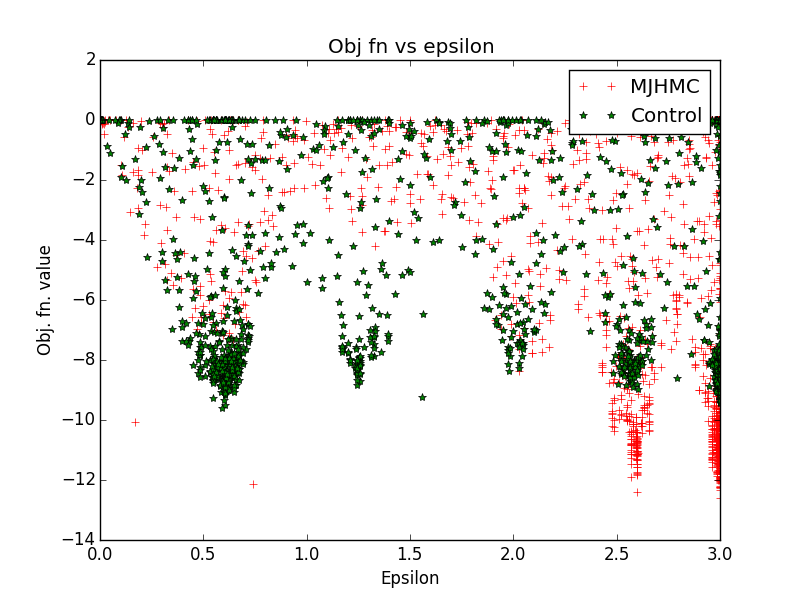}
\caption{Search performance projected onto the $\eps$ axis.
\label{fig search eps}}
\end{figure}

\begin{figure}[H]
\centering
\includegraphics[width=.75\textwidth]{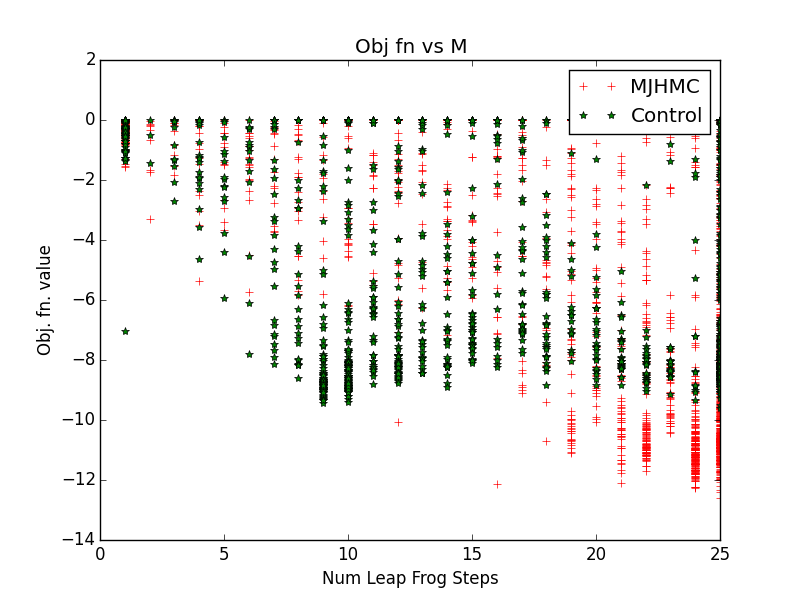}
\caption{Search performance projected onto the $M$ axis.
\label{fig search M}}
\end{figure}

Figures \ref{fig search beta}, \ref{fig search eps}, \ref{fig search M} illustrate the overall structure of Spearmint's search for hyperparameters. The green stars represent a trial setting
of MJHMC hyperpameter and the red crosses represent a trial setting of standard HMC hyperparameters.
The y-axis represents the value of the objective function for each trial setting. It can be seen that in \ref{fig search beta} that MJHMC chooses smaller $\beta$ values which suggests wanting to corrupt momentum more slowly as compared to the control case. It can also be seen from \ref{fig search eps,fig search M} that MJHMC prefers larger steplengths for the integrator ($\epsilon$) and steps ($M$) . 

\section{Derivation of equation \ref{eq pbar relation}}

First we calculate $P(\tau_2 \leq \tau_1)$ where $\tau_i$ is drawn from $\text{Exp}(\lambda_i)$ for $i = 1,2$:

\begin{align*}
P(\tau_2 \leq \tau_1) &= \int_0^\infty P(\tau_{1} = tdt)P(\tau_2 \leq t) dt\\
&= \int_0^\infty  (\lambda_1 \exp(-\lambda_1 t))  \int_0^t(\lambda_2 \exp(-\lambda_2 \tau)) d\tau dt \\
&= \lambda_1  \int_0^\infty \exp(-\lambda_1 t) \left[1 - \exp(-\lambda_2 t)\right] dt \\
&= \lambda_1 \int_0^\infty \exp(-\lambda_1 t) - \exp(- (\lambda_1 + \lambda_2) t) dt\\
&= \lambda_1  \left[\left. -\frac{1}{\lambda_1}  \exp(-\lambda_1 t) \right\rvert_0^\infty \right]
- \left[\left.- \frac{1}{\lambda_1 + \lambda_2} \exp(-(\lambda_1 + \lambda_1) t) \right\rvert_0^\infty \right] \\
&= \lambda_1 \left[ \frac{1}{\lambda_1} - \frac{1}{\lambda_1 + \lambda_2}\right] \\
&=  1 - \frac{\lambda_1}{\lambda_1 + \lambda_2}  \\
&= \frac{\lambda_2}{\lambda_1 + \lambda_2}
\end{align*}

Let $\tau_{i,j}$ be drawn from $\text{Exp}(\Gamma_{ij})$. Then

\begin{align*}
P(\zeta_j \mid \zeta_i) 
&=P(\tau_{i,j} = \min\set{\tau_{i,1}, \tau_{i,2}, \dots \tau_{i,n}}) \\
&= \prod_{k \neq j}^n  P(\tau_{i,j} \leq \tau_{i,k}) \\
&= \prod_{k \neq j}^n \frac{\Gamma_{i,j}}{\Gamma_{i,j} + \Gamma_{i,k}}
\end{align*}

\end{document}